%% file: iclr2024_conference.tex
\documentclass{article} 
\usepackage{iclr2024_conference,times}

\input{math_commands.tex}

\usepackage{hyperref}
\usepackage{url}

\usepackage{algorithm}
\usepackage{algorithmic}
\usepackage{adjustbox}
\usepackage{multirow}
\usepackage{xcolor}
\usepackage{xcolor, soul}
\usepackage{amssymb}
\usepackage{amsmath}
\usepackage[utf8]{inputenc}
\usepackage{kotex}
\usepackage{bm}
\usepackage{wrapfig,lipsum,booktabs}
\usepackage{stfloats}
\usepackage[labelformat=empty]{subfig}

\title{SPI-GAN: Denoising Diffusion GANs with \\ Straight-Path Interpolations}

\author{Jinsung Jeon \\
Yonsei University\\
\texttt{jjsjjs0902@yonsei.ac.kr} \\
\And
Noseong Park\\
KAIST \\
\texttt{noseong@kaist.ac.kr} \\
}

\iclrfinalcopy 

\begin{document}

\maketitle


\begin{abstract}
Score-based generative models (SGMs) show the state-of-the-art sampling quality and diversity. However, their training/sampling complexity is notoriously high due to the highly complicated forward/reverse processes, so they are not suitable for resource-limited settings. To solving this problem, learning a simpler process is gathering much attention currently. We present an enhanced GAN-based denoising method, called SPI-GAN, using our proposed \emph{straight-path interpolation} definition. To this end, we propose a GAN architecture i) denoising through the straight-path and ii) characterized by a continuous mapping neural network for imitating the denoising path. This approach drastically reduces the sampling time while achieving as high sampling quality and diversity as SGMs. As a result, SPI-GAN is one of the best-balanced models among the sampling quality, diversity, and time for CIFAR-10, and CelebA-HQ-256.
\end{abstract}

\section{Introduction}
\label{introduction}
Generative models are one of the most popular research topics for deep learning. 
In particular, the diffusion models~\citep{song2019generative,ho2020denoising}, which progressively corrupt original data and revert the corruption process to generate an image, have recently shown good performance.
%
%
%
\citet{song2020score} proposed a stochastic differential equation (SDE)-based mechanism that embraces all those models and coined the term, \emph{score-based generative models} (SGMs).

Although SGMs show good performance in terms of sampling quality and sampling diversity, they take a lot of time for sampling, which is one of the generation task trilemmas.
This problem becomes more problematic in resource-limited environments, and for this reason, reducing the complexity of SGM during sampling is gathering much attention. 
There exist two different directions for this: i) learning a simpler process than the complicated forward/reverse process of SGMs~\citep{das2023image}, and ii) letting GANs imitate
\footnote{Let $\mathbf{x}_0$ be a clean original sample 
and $\mathbf{x}_T$ be a noisy sample that follows a Gaussian prior 
under the context of SGMs. These imitation methods learn a denoising process following the reverse SDE path by training their conditional generators to read $\mathbf{x}_t$ and output $\mathbf{x}_{t-j}$. Typically, a large $j > 0$ is preferred to reduce the denoising step (see Section~\ref{relatedwork} for a detailed explanation).}
SGMs~\citep{xiao2021tackling,wang2022diffusion}. 

\begin{figure*}[t]
\vspace{-2em}
\centering
        \subfloat[(a) The SDE-based workflow of SGMs. The reverse SDE is a generation process.]{\includegraphics[width=0.24\textwidth]{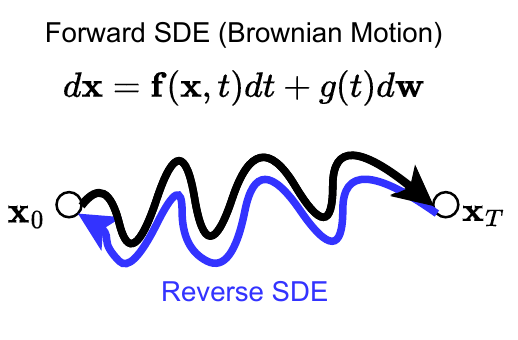}}\hfill
        \subfloat[(b) Approximating the reverse SDE with $K$ shortcuts by DD-GAN, e.g., $K=3$ in this example.]{\includegraphics[width=0.24\textwidth]{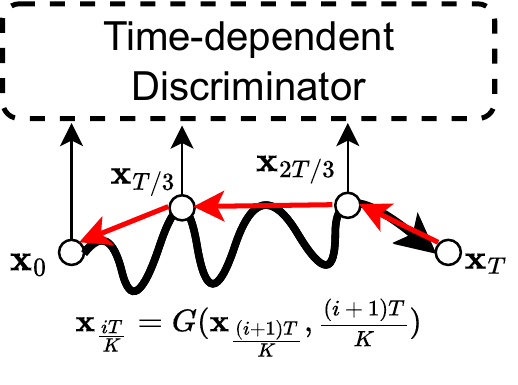}}\hfill
        \subfloat[(c) Augmentation following the forward SDE of Diffusion-GAN.]{\includegraphics[width=0.24\textwidth]{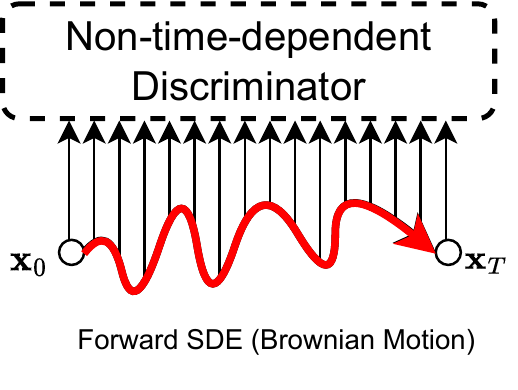}}\hfill
        \subfloat[(d) Our proposed denoising method, SPI-GAN, with straight-path interpolations.]{\includegraphics[width=0.24\textwidth]{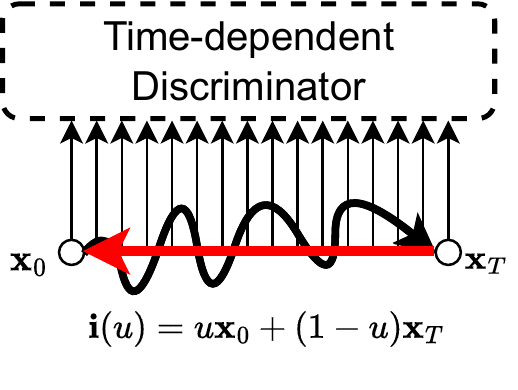}}
    \caption{The comparison among four models: i) the original formulation of SGMs in (a), ii) DD-GAN's learning the shortcuts of the reverse SDE in (b), iii) Diffusion-GAN's augmentation method in (c) and iv) SPI-GAN's learning the straight path in (d). The red paths in (b), (c), and (d) are used as the discriminators' input.} 
    \label{fig:arch1}
    \vspace{-1em}
\end{figure*}

%
Our method can be considered a hybrid of the previous works. 
%
Inspired by them, we propose a GAN-base hybrid method that approximates the linear interpolation. 
Our method imitates a denoising process following the straight-path interpolation guided by $u\mathbf{x}_0+(1-u)\mathbf{x}_T$, where $0 \leq u \leq 1$ (cf. Figure~\ref{fig:arch1} (d)). 
Therefore, we call our method \emph{straight-path interpolation GAN} (SPI-GAN).

One may consider that SPI-GAN is similar to Denoising Diffusion GAN (DD-GAN) and Diffusion-GAN (cf. Figure~\ref{fig:arch1} (b-d)). 
DD-GAN~\citep{xiao2021tackling} effectively reduces the number of denoising steps by letting its GAN approximate the shortcuts. 
In Diffusion-GAN~\citep{wang2022diffusion}, image augmentation is performed by injecting noisy images following the forward path of SGMs, and the standard adversarial training is conducted for its GAN. 
However, our SPI-GAN is technically different and more sophisticated in the following points:
\begin{itemize}
    \item Our straight-path interpolation is as simple as Eq.~\ref{eq:interpolation}, is much easier to learn.
    It is also possible to derive its ordinary differential equation (ODE)-based formulation, which we call the straight-path interpolation process. 
    %
    \item In DD-GAN, the generator learns $K$ shortcuts through the reverse path of SGMs, whereas SPI-GAN learns the straight-interpolation path.
    \item Diffusion-GAN's discriminator is trained with the augmented noisy images without being conditioned on time --- in other words, it is non-time-dependent. 
    SPI-GAN's time-dependent discriminator learns the straight-path information, being conditioned on time.
    \item In order to learn a simple process, i.e., our straight-path interpolation, SPI-GAN uses a special neural network architecture, characterized by a mapping network.
    \item After all these efforts, SPI-GAN is a GAN-based method that imitates the straight-path interpolation. 
    However, our mapping network is designed for this purpose, allowing SPI-GAN to generate fake images directly without recursion.
\end{itemize}

DD-GAN, Diffusion-GAN, and SPI-GAN attempt to address the trilemma of the generative model using GANs. 
Among those models, our proposed SPI-GAN, which learns a straight-path interpolation, shows the best balance in terms of the overall sampling quality, diversity, and time in two different resolution benchmark datasets: CIFAR-10, and CelebA-HQ-256

\section{Related work and preliminaries}
\label{relatedwork}

\paragraph{Neural ordinary differential equations (NODEs).} Neural ordinary differential equations~\citep{chen2018neural} use the following equation to define the continuous evolving process of the hidden vector $\mathbf{h}(u)$:
\begin{align}\label{eq:node}
\mathbf{h}(u) = \mathbf{h}(0) + \int_{0}^{u} f(\mathbf{h}(t), t; \boldsymbol{\theta}_\mathbf{f}) dt,
\end{align}
where the neural network $f(\mathbf{h}(t), t;\boldsymbol{\theta}_\mathbf{f})$ learns $\frac{d \mathbf{h}(t)}{dt}$. 
To solve the integral problem, we typically rely on various ODE solvers. 
The explicit Euler method is one of the simplest ODE solvers. 
The 4th order Runge--Kutta (RK4) method is a more sophisticated ODE solver and the Dormand--Prince~\citep{DORMAND198019} method is an advanced adaptive step-size solver. 
The NODE-based continuous-time models~\citep{chen2018neural,DBLP:conf/nips/KidgerMFL20} show good performance in processing sequential data.

\paragraph{Letting GANs imitate SGMs.} Among the ways to solve the SGMs trilemma, letting GANs imitate SGMs shows the best-balanced performance in terms of the generative task trilemma. 
Figure~\ref{fig:arch1} (b) shows the key idea of DD-GAN, which proposed to approximate the reverse SDE process with $K$ shortcuts. 
They internally utilize a GAN-based framework conditioned on time (step) $t$ to learn the shortcuts. 
A generator of DD-GAN receives noise images as input and denoises it. 
In other words, the generator generates clean images through denoising $K$ times from the noise images. 
To this end, DD-GAN uses a conditional generator, and $\mathbf{x}_{t}$ is used as a condition to generate $\mathbf{x}_{t-1}$. 
It repeats this $K$ times to finally creates $\mathbf{x}_{0}$. 
For its adversarial learning, the generator can match $p_{\boldsymbol{\theta}}(\mathbf{x}_{t-1}|\mathbf{x}_{t})$ and $q(\mathbf{x}_{t-1}|\mathbf{x}_{t})$. 
Figure~\ref{fig:arch1} (c) shows the overall method of Diffusion-GAN. 
In contrast to DD-GAN, Diffusion-GAN directly generates clean images and its discriminator is trained after augmenting real images with noisy images that follow the forward SDE from the generated images. 
The noisy images are used as input to the discriminator to prevent mode-collapse in GANs.

\begin{figure*}[t]
    \begin{center}    {\includegraphics[width=0.8\textwidth,trim={0 0 0 0},clip]{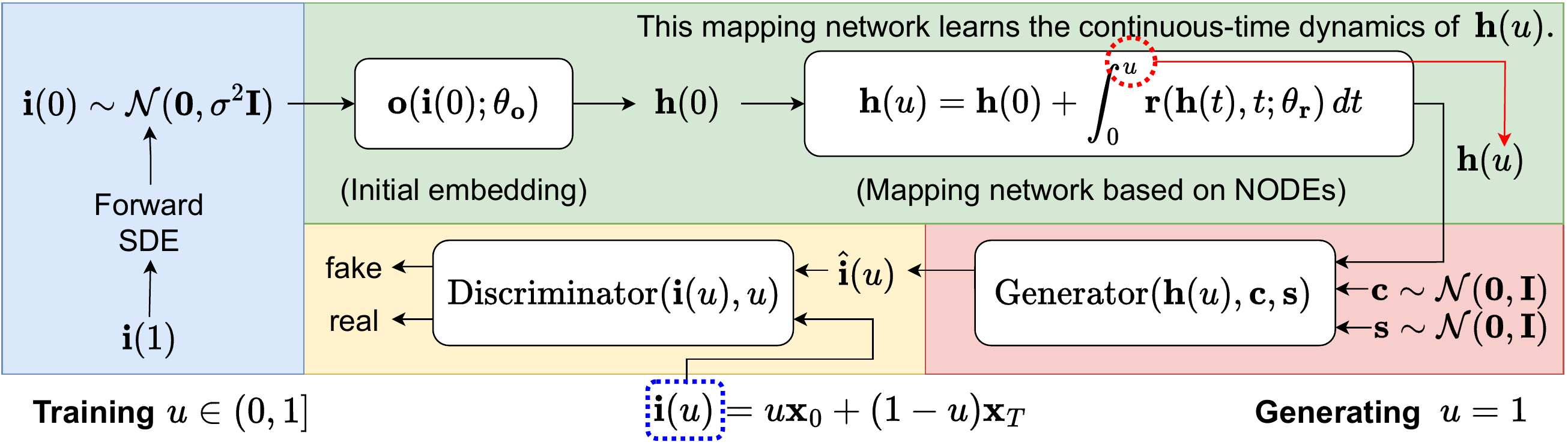}}
    \caption{The architecture of our proposed SPI-GAN. $\textbf{h}(u)$ is a latent vector which generates an interpolated image $\textbf{i}(u)$ at time $u$. Therefore, $\mathbf{i}(1)$ is an original image and $\mathbf{i}(0)$ is a noisy image. We perform this adversarial training every time $u$ but generate images with $u=1$. The constant noise $\mathbf{c}$ and the layer-wise varying noise $\mathbf{s}$ enable the stochasticity of the generator.} 
    \label{fig:framework}
    \end{center}
    \vspace{-1em}
\end{figure*}

\section{Proposed method}
\label{proposedmethod} 
%

%
Before describing our proposed method, the notations in this paper are defined as follows: 
i) $\mathbf{i}(u) \in \mathbb{R}^{C \times H \times W}$ is an image with a channel $C$, a height $H$, and a width $W$ at interpolation point $u$. 
ii) $\hat{\mathbf{i}}(u) \in \mathbb{R}^{C \times H \times W}$ is a generated fake image at interpolation point $u$. 
iii) $\mathbf{h}(u) \in \mathbb{R}^{\dim(\mathbf{h})}$ is a latent vector at interpolation point $u$. 
iv) $\mathbf{r}(\mathbf{h}(t), t; \mathbf{\boldsymbol{\theta}}_\mathbf{r})  \in \mathbb{R}^{\dim(\mathbf{h})}$ is a neural network approximating the time derivative of $\mathbf{h}(t)$, denoted $\frac{d\mathbf{h}(t)}{dt}$.
Our proposed SPI-GAN consists of four parts, 
each of which has a different color in Figure~\ref{fig:framework} 

\subsection{Diffusion through the forward SDE}
In the first part highlighted in blue, $\sigma^2$ can be different for $\mathbf{i}(0) \sim \mathcal{N}(\mathbf{0}, \sigma^{2}\mathbf{I})$ depending on the type of the forward SDE. 
%
%
Unlike the reverse SDE, which requires step-by-step computation, the forward SDE can be calculated with one-time computation for a target time $t$~\citep{song2020score}. 
Therefore, it takes $\mathcal{O}(1)$ in the first part. 
We note that $\mathbf{i}(0) = \mathbf{x}_T$ and $\mathbf{i}(1) = \mathbf{x}_0$ are the two ends of the straight-path interpolation which will be described shortly.

\subsection{Straight-path interpolation}
SPI-GAN has advantages over SGMs since it learns a much simpler process than the SDE-based path of SGMs. 
Our straight-path interpolation (SPI) is defined as follows:
\begin{align}\label{eq:interpolation}
\mathbf{i}(u) = u\mathbf{x}_0 + (1-u)\mathbf{x}_T,
\end{align}
where $u \in (0,1]$ is an intermediate point and therefore, $\mathbf{i}(1) = \mathbf{x}_0$ and $\mathbf{i}(0) = \mathbf{x}_T$. 
We use a straight-path between $\mathbf{i}(1)$ and $\mathbf{i}(0)$ to learn the shortest path with the minimum Wasserstein distance. 
Therefore, Our straight-path interpolation process is simpler and thus more suitable for neural networks to learn. 
During our training process, we randomly sample $u$ every iteration. 
%

\subsection{Mapping network}
\begin{figure*}[t]
    \centering
    \begin{minipage}{0.45\linewidth}
        \centering
        \includegraphics[width=0.8\linewidth]{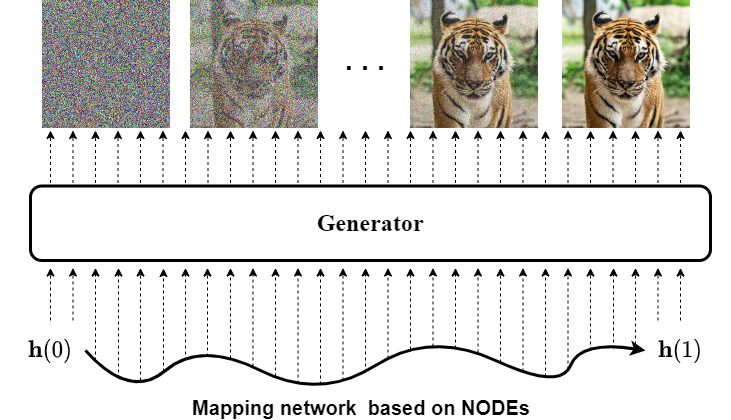}
        \caption{Process of generating sample from continuous latent vector.}
        \label{fig:detailmapping}
    \end{minipage}
    \hspace{0.05\linewidth}
    \begin{minipage}{0.45\linewidth}
        \centering
        \includegraphics[width=0.8\linewidth]{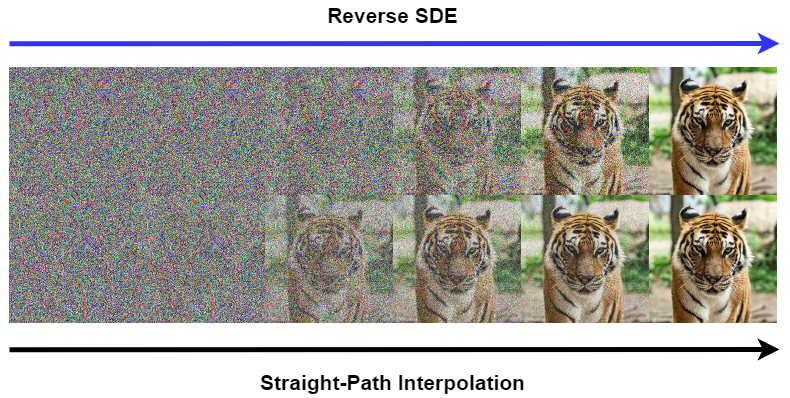}
        \caption{Difference between reverse SDE and interpolation.}
        \label{fig:sdevsspi}
    \end{minipage}
\end{figure*}

The mapping network, which generates a latent vector $\mathbf{h}(u)$, is the most important component in our model. 
The mapping network consists of an initial embedding network, denoted $\mathbf{o}$, that generates the initial hidden representation $\mathbf{h}(0)$ from $\mathbf{i}(0)$, and a NODE-based mapping network. 
The role of network $\mathbf{o}$ is to reduce the size of the input to the mapping network for decreasing sampling time. 
In addition, the NODE-based mapping network generates the latent vector for a target interpolation point $u$, whose initial value problem (IVP) is defined as follows:
\begin{align}\begin{split}\label{eq:mapping}
 {\mathbf{h}}\big(u) &= \mathbf{h}(0) + \int_{0}^{u} \mathbf{r}(\mathbf{h}(t), t;\mathbf{\boldsymbol{\theta}}_\mathbf{r}) dt,
\end{split}\end{align}
where $\frac{d\mathbf{h}(t)}{dt} = \mathbf{r}(\mathbf{h}(t), t;\mathbf{\boldsymbol{\theta}}_\mathbf{r})$, and $\mathbf{r}$ has multiple fully-connected layers in our implementation. 
In general, $\mathbf{h}(0)$ is a lower-dimensional representation of the input. 
One more important point is that we maintain a single latent space for all $u$ and therefore, $\mathbf{h}(u)$ has the information of the image to generate at a target interpolation time $u$. 
For instance, Figure~\ref{fig:detailmapping} shows that a noisy image is generated from $\mathbf{h}(0)$ but a clean image from $\mathbf{h}(1)$.
%


\subsection{Generator} We customize the generator architecture of StyleGAN2~\citep{karras2020analyzing} for our purposes. 
Our generator takes $\mathbf{c}, \mathbf{s}$ as input and mimics the stochastic property of SGMs. 
However, the biggest difference from StyleGAN2 is that our generator is trained by the \emph{continuous-time} mapping network which generates latent vectors at various interpolation points while maintaining one latent space across them. 
This is the key point in our model design to generate the interpolated image $\hat{\mathbf{i}}(u)$ with various $u$ settings. 

\subsection{Discriminator} The discriminator of SPI-GAN is time-dependent, unlike the discriminator of traditional GANs. 
It takes $\hat{\mathbf{i}}(u)$ and the embedding of $u$ as input and learns to classify images from various interpolation points. 
As a result, it i) learns the data distribution which changes following the straight-path interpolation, and ii) solves the mode-collapse problem that traditional GANs have, since our discriminator sees various clean and noisy images. 
DD-GAN and Diffusion-GAN also use this strategy to overcome the mode-collapse problem. 
They give noise images following the reverse (forward) SDE path as input to the discriminator. 
As shown in Figure~\ref{fig:sdevsspi}, however, the straight-path interpolation method 
maintains a better balance between noisy and clean images than the case where we sample following the SDE path. 
Therefore, our discriminator learns a more balanced set of noisy and clean images than that of DD-GAN and Diffusion-GAN. 
%



\subsection{How to generate}
\label{sec:gen}
In order to generate samples with SPI-GAN, we need only $\mathbf{h}(1)$ from the mapping network. 
Unlike other auto-regressive denoising models that require multiple steps when generating samples, e.g., DD-GAN (cf. Figure~\ref{fig:arch1} (b)), SPI-GAN learns the denoising path using a NODE-based mapping network. 
After sampling $\mathbf{z} \sim \mathcal{N}(\mathbf{0}, \sigma^{2}\mathbf{I})$, we feed them into the mapping network to derive $\mathbf{h}(1)$ --- we solve the initial value problem in the Eq.~\ref{eq:mapping} from 0 to 1 ---  and our generator generates a fake image $\hat{\mathbf{i}}(1)$. 
In other words, it is possible to generate latent vector $\mathbf{h}(1)$ directly in our case, which is later used to generate a fake sample $\hat{\mathbf{i}}(1)$. 

\section{Experiments}
\label{experiments}

For evaluating our proposed method, SPI-GAN, we use different resolution benchmark datasets. CIFAR-10~\citep{krizhevsky2014cifar} has a resolution of 32x32. CelebA-HQ-256~\citep{karras2017progressive} contains high-resolution images of 256x256. We use 5 evaluation metrics to quantitatively evaluate fake images. The inception score~\citep{salimans2016improved} and the Fr\'echet inception distance~\citep{heusel2017gans} are traditional methods to evaluate the fidelity of fake samples. The improved recall~\citep{kynkaanniemi2019improved} reflects whether the variation of generated data matches the variation of training data. Finally, the number of function evaluations (NFE).  

\begin{table*}[t]
\centering
\begin{minipage}{.6\textwidth}
\caption{Results of the generation on CIFAR-10.}
\begin{adjustbox}{width=\textwidth}
\begin{tabular}{lrrrr}
\hline
Model & IS $\uparrow$ &FID $\downarrow$ & Recall$\uparrow$ & NFE$\downarrow$  \\ \hline
SPI-GAN (ours) & \textbf{10.2} & 3.01 & \textbf{0.66} & 1  \\ \hline
Diffusion-GAN (StyleGAN2)~\citep{wang2022diffusion} & 9.94 & 3.19 & 0.58 & 1  \\
Denoising Diffusion GAN (DD-GAN), $K=4$~\citep{xiao2021tackling} & 9.63 & 3.75 & 0.57 & 4  \\ \hline
Score SDE (VP)~\citep{song2020score} & 9.68 & 2.41 & 0.59 & 2000  \\ 
DDPM~\citep{ho2020denoising} & 9.46 & 3.21 & 0.57 & 1000  \\
NCSN~\citep{song2019generative} & 8.87 & 25.3 & - & 1000  \\
Score SDE (VE)~\citep{song2020score} & 9.89 & 2.20 & 0.59 & 2000  \\
LSGM~\citep{vahdat2021score} & 9.87 & \textbf{2.10} & 0.61 & 147 \\
DDIM, T=50~\citep{song2020denoising} & 8.78 & 4.67 & 0.53 & 50  \\
FastDDPM, T=50~\citep{kong2021fast} & 8.98 & 3.41 & 0.56 & 50 \\
Recovery EBM~\citep{gao2020learning} & 8.30 & 9.58 & - & 180  \\
Improved DDPM~\citep{nichol2021improved} & - & 2.90 & - & 4000  \\ \hline
StyleGAN2 w/o ADA~\citep{karras2020analyzing} & 9.18 & 8.32 & 0.41 & 1  \\ 
StyleGAN2 w/ ADA~\citep{karras2020training} & 9.83 & 2.92 & 0.49 & 1  \\ 
StyleGAN2 w/ Diffaug~\citep{zhao2020differentiable} & 9.40 & 5.79 & 0.42 & 1  \\ \hline
\end{tabular}
\end{adjustbox}
\label{tbl:cifar10}
\end{minipage}
\begin{minipage}{.39\textwidth}
\begin{minipage}{\textwidth}
\vspace{1em}
\caption{Results on CelebA-HQ-256.}\label{tbl:celeba}
\vspace{-1em}
\centering
\scriptsize
\begin{center}
\begin{small}
\begin{sc}
\begin{adjustbox}{width=0.75\textwidth}
\begin{tabular}{lr}\hline
Model &  FID$\downarrow$  \\  \hline
SPI-GAN (ours) & \textbf{6.62} \\ \hline
DD-GAN & 7.64 \\ 
Score SDE & 7.23   \\ \hline
LSGM & 7.22  \\
UDM & 7.16  \\ \hline
PGGAN~\citep{karras2017progressive} & 8.03 \\
Adv. LA~\citep{pidhorskyi2020adversarial} & 19.2\\
VQ-GAN~\citep{esser2021taming} & 10.2\\
DC-AE~\citep{parmar2021dual} &  15.8 \\ \hline
\end{tabular}
\end{adjustbox}
\end{sc}
\end{small}
\end{center}
\end{minipage}
\end{minipage}
\vspace{-1em}
\end{table*}


\begin{wrapfigure}{r}{5cm}
{\includegraphics[width=0.36\textwidth]{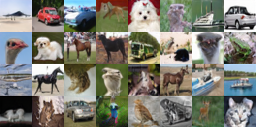}
     \includegraphics[width=0.36\textwidth]{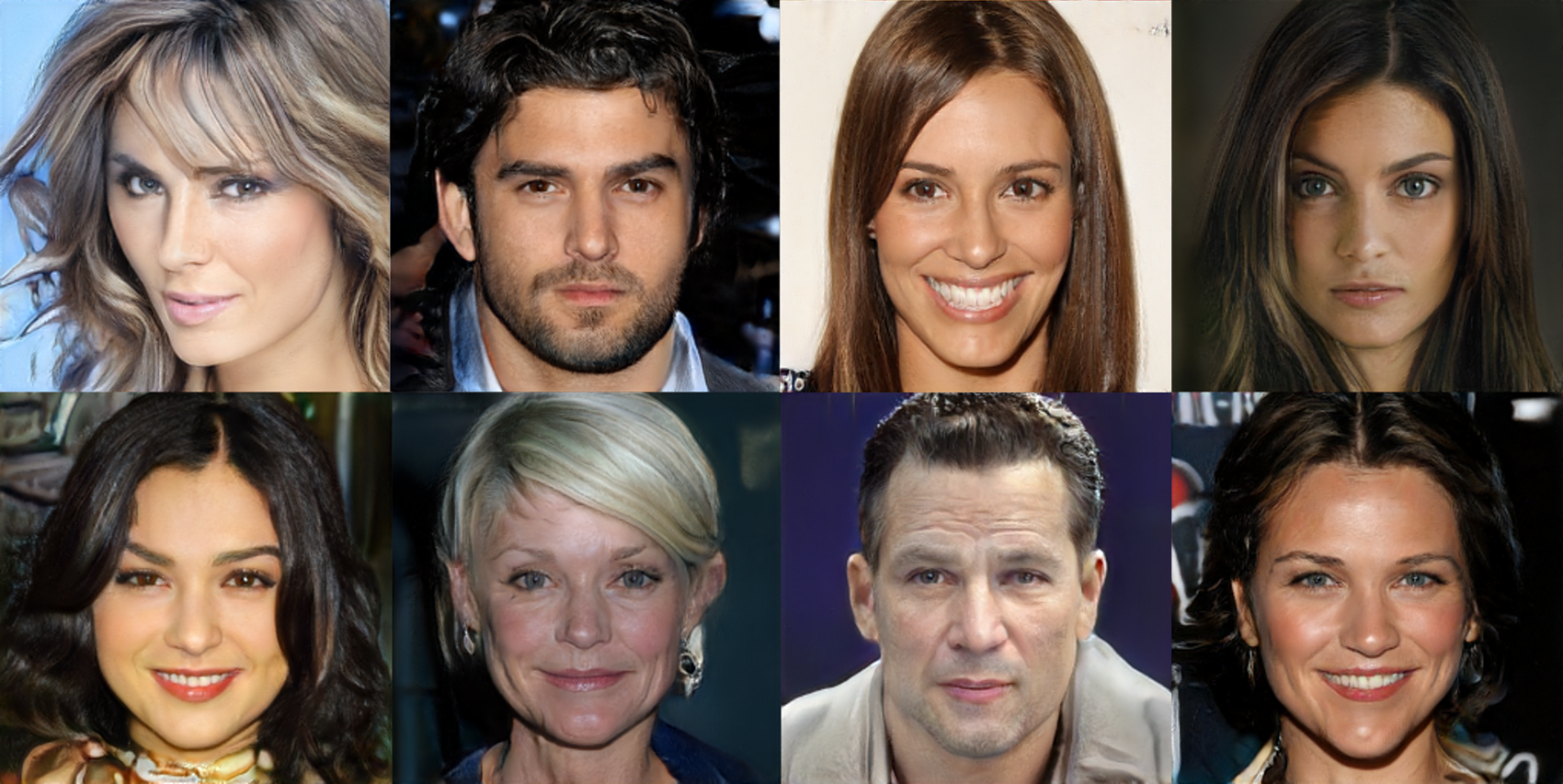}}
    \caption{Qualitative results on CIFAR-10, and CelebA-HQ-255.}
    \label{fig:lsun}
    \vspace{-2em}
\end{wrapfigure}
\subsection{Main results}
\label{mainresults}
In this subsection, we evaluate our proposed model quantitatively and qualitatively. 
For CIFAR-10, we perform the unconditional image generation task for fair comparisons with existing models. 
The quantitative evaluation results are shown in Table~\ref{tbl:cifar10}. 
Although our Fr\'echet inception distance (FID) is 0.6 worse than that of the Score SDE (VP), it shows better scores in all other metrics. However, our method has a better FID score than that of DD-GAN and Diffusion-GAN, which are the most related methods. 
Diffusion-GAN, which follows the reverse SDE path using GAN, is inferior to our method for all those three quality metrics. 
LGSM also shows high quality for FID. However, its inception score (IS) and improved recall (Recall) scores are worse than ours.

Even for high-resolution images, our model shows good performance. 
In particular, our method shows the best FID score for CelebA-HQ-256 in Table~\ref{tbl:celeba}, which shows the efficacy of our proposed method. 
As shown, our method is able to generate visually high-quality images.

\subsection{Additional studies}

\paragraph{Sampling time analyses.}
\label{trainingtimeanalyses}

\begin{wraptable}{r}{6.5cm}
\vspace{-1.3em}
\caption{The Generation time comparison}\label{tbl:time}
\begin{center}
\begin{adjustbox}{width=0.35\textwidth}
\begin{tabular}{lr}
\hline
Model & Time \\
\hline
SPI-GAN (Ours)   & 0.04  \\
\hline
\begin{tabular}[l]{@{}l@{}}Diffusion-GAN (StyleGAN2)\end{tabular} & 0.04  \\
DD-GAN &  0.36 \\
StyleGAN2    & 0.04  \\
\hline
\end{tabular}
\end{adjustbox}
\end{center}
\vspace{-1em}
\end{wraptable}
Our model shows outstanding performance in all evaluation metrics compared to DD-GAN and Diffusion-GAN. 
In particular, SPI-GAN, unlike DD-GAN, does not increase the sample generation time --- in fact, our method only affects the training time because we train our method with $u \in (0,1]$. 
However, we always use $u=1$ for generating a clean image $\hat{\mathbf{i}}(1)$. 
Therefore, our method is fast in generating images after being trained, which is one good characteristic of our method. 
We measure the wall-clock runtime 10 times for CIFAR-10 with a batch size of 100 using an A5000 GPU to evaluate the sampling time. 
We applied the same environment for fair comparisons. 
As a result, our method's sample generation time in Table~\ref{tbl:time} is almost the same as that of StyleGAN2, which is one of the fastest methods. 
In summary, SPI-GAN not only increases the quality of samples but also decreases the sampling time.

\section{Conclusions and discussions}
\label{conclusions}
Score-based generative models (SGMs) now show the state-of-the-art performance in image generation. 
However, the sampling time of SGMs is significantly longer than other generative models, such as GANs, VAEs, and so on. 
Therefore, we presented the most balanced model by reducing the sampling time, called SPI-GAN. 
Our method is a GAN-based approach that imitates the straight-path interpolation. 
%
%
Moreover, it can directly generate fake samples without any recursive computation (or step). 
Our method shows the best sampling quality in various metrics and faster sampling time than other score-based methods. 
%
%
%
%
All in all, one can see that SPI-GAN is one of the most balanced methods among the generative task trilemma's criteria: sampling quality, diversity, and time. 

\clearpage

\bibliography{iclr2024_conference}
\bibliographystyle{iclr2024_conference}
\clearpage

\end{document}

%% file: math_commands.tex
\usepackage{amsmath,amsfonts,bm}

\def\eqref#1{equation~\ref{#1}}

\def\1{\bm{1}}

\DeclareMathAlphabet{\mathsfit}{\encodingdefault}{\sfdefault}{m}{sl}
\SetMathAlphabet{\mathsfit}{bold}{\encodingdefault}{\sfdefault}{bx}{n}













%% file: iclr2024_conference.bbl
\begin{thebibliography}{25}
\providecommand{\natexlab}[1]{#1}
\providecommand{\url}[1]{\texttt{#1}}
\expandafter\ifx\csname urlstyle\endcsname\relax
  \providecommand{\doi}[1]{doi: #1}\else
  \providecommand{\doi}{doi: \begingroup \urlstyle{rm}\Url}\fi

\bibitem[Chen et~al.(2018)Chen, Rubanova, Bettencourt, and Duvenaud]{chen2018neural}
Ricky~TQ Chen, Yulia Rubanova, Jesse Bettencourt, and David~K Duvenaud.
\newblock Neural ordinary differential equations.
\newblock \emph{Advances in neural information processing systems}, 31, 2018.

\bibitem[Das et~al.(2023)Das, Fotiadis, Batra, Nabiei, Liao, Vakili, Shiu, and Bernacchia]{das2023image}
Ayan Das, Stathi Fotiadis, Anil Batra, Farhang Nabiei, FengTing Liao, Sattar Vakili, Da-shan Shiu, and Alberto Bernacchia.
\newblock Image generation with shortest path diffusion.
\newblock \emph{arXiv preprint arXiv:2306.00501}, 2023.

\bibitem[Dormand \& Prince(1980)Dormand and Prince]{DORMAND198019}
J.R. Dormand and P.J. Prince.
\newblock A family of embedded runge-kutta formulae.
\newblock \emph{Journal of Computational and Applied Mathematics}, 6\penalty0 (1):\penalty0 19 -- 26, 1980.

\bibitem[Esser et~al.(2021)Esser, Rombach, and Ommer]{esser2021taming}
Patrick Esser, Robin Rombach, and Bjorn Ommer.
\newblock Taming transformers for high-resolution image synthesis.
\newblock In \emph{Proceedings of the IEEE/CVF Conference on Computer Vision and Pattern Recognition}, pp.\  12873--12883, 2021.

\bibitem[Gao et~al.(2021)Gao, Song, Poole, Wu, and Kingma]{gao2020learning}
Ruiqi Gao, Yang Song, Ben Poole, Ying~Nian Wu, and Diederik~P Kingma.
\newblock Learning energy-based models by diffusion recovery likelihood.
\newblock \emph{arXiv preprint arXiv:2012.08125}, 2021.

\bibitem[Heusel et~al.(2017)Heusel, Ramsauer, Unterthiner, Nessler, and Hochreiter]{heusel2017gans}
Martin Heusel, Hubert Ramsauer, Thomas Unterthiner, Bernhard Nessler, and Sepp Hochreiter.
\newblock Gans trained by a two time-scale update rule converge to a local nash equilibrium.
\newblock \emph{Advances in neural information processing systems}, 30, 2017.

\bibitem[Ho et~al.(2020)Ho, Jain, and Abbeel]{ho2020denoising}
Jonathan Ho, Ajay Jain, and Pieter Abbeel.
\newblock Denoising diffusion probabilistic models.
\newblock \emph{Advances in Neural Information Processing Systems}, 33:\penalty0 6840--6851, 2020.

\bibitem[Karras et~al.(2018)Karras, Aila, Laine, and Lehtinen]{karras2017progressive}
Tero Karras, Timo Aila, Samuli Laine, and Jaakko Lehtinen.
\newblock Progressive growing of gans for improved quality, stability, and variation.
\newblock \emph{arXiv preprint arXiv:1710.10196}, 2018.

\bibitem[Karras et~al.(2020{\natexlab{a}})Karras, Aittala, Hellsten, Laine, Lehtinen, and Aila]{karras2020training}
Tero Karras, Miika Aittala, Janne Hellsten, Samuli Laine, Jaakko Lehtinen, and Timo Aila.
\newblock Training generative adversarial networks with limited data.
\newblock \emph{Advances in Neural Information Processing Systems}, 33:\penalty0 12104--12114, 2020{\natexlab{a}}.

\bibitem[Karras et~al.(2020{\natexlab{b}})Karras, Laine, Aittala, Hellsten, Lehtinen, and Aila]{karras2020analyzing}
Tero Karras, Samuli Laine, Miika Aittala, Janne Hellsten, Jaakko Lehtinen, and Timo Aila.
\newblock Analyzing and improving the image quality of stylegan.
\newblock In \emph{Proceedings of the IEEE/CVF conference on computer vision and pattern recognition}, pp.\  8110--8119, 2020{\natexlab{b}}.

\bibitem[Kidger et~al.(2020)Kidger, Morrill, Foster, and Lyons]{DBLP:conf/nips/KidgerMFL20}
Patrick Kidger, James Morrill, James Foster, and Terry~J. Lyons.
\newblock Neural controlled differential equations for irregular time series.
\newblock In \emph{NeurIPS}, 2020.

\bibitem[Kong \& Ping(2021)Kong and Ping]{kong2021fast}
Zhifeng Kong and Wei Ping.
\newblock On fast sampling of diffusion probabilistic models.
\newblock \emph{arXiv preprint arXiv:2106.00132}, 2021.

\bibitem[Krizhevsky et~al.(2014)Krizhevsky, Nair, and Hinton]{krizhevsky2014cifar}
Alex Krizhevsky, Vinod Nair, and Geoffrey Hinton.
\newblock The cifar-10 dataset.
\newblock \emph{online: http://www. cs. toronto. edu/kriz/cifar. html}, 55\penalty0 (5), 2014.

\bibitem[Kynk{\"a}{\"a}nniemi et~al.(2019)Kynk{\"a}{\"a}nniemi, Karras, Laine, Lehtinen, and Aila]{kynkaanniemi2019improved}
Tuomas Kynk{\"a}{\"a}nniemi, Tero Karras, Samuli Laine, Jaakko Lehtinen, and Timo Aila.
\newblock Improved precision and recall metric for assessing generative models.
\newblock \emph{Advances in Neural Information Processing Systems}, 32, 2019.

\bibitem[Nichol \& Dhariwal(2021)Nichol and Dhariwal]{nichol2021improved}
Alexander~Quinn Nichol and Prafulla Dhariwal.
\newblock Improved denoising diffusion probabilistic models.
\newblock In \emph{International Conference on Machine Learning}, pp.\  8162--8171. PMLR, 2021.

\bibitem[Parmar et~al.(2021)Parmar, Li, Lee, and Tu]{parmar2021dual}
Gaurav Parmar, Dacheng Li, Kwonjoon Lee, and Zhuowen Tu.
\newblock Dual contradistinctive generative autoencoder.
\newblock In \emph{Proceedings of the IEEE/CVF Conference on Computer Vision and Pattern Recognition}, pp.\  823--832, 2021.

\bibitem[Pidhorskyi et~al.(2020)Pidhorskyi, Adjeroh, and Doretto]{pidhorskyi2020adversarial}
Stanislav Pidhorskyi, Donald~A Adjeroh, and Gianfranco Doretto.
\newblock Adversarial latent autoencoders.
\newblock In \emph{Proceedings of the IEEE/CVF Conference on Computer Vision and Pattern Recognition}, pp.\  14104--14113, 2020.

\bibitem[Salimans et~al.(2016)Salimans, Goodfellow, Zaremba, Cheung, Radford, and Chen]{salimans2016improved}
Tim Salimans, Ian Goodfellow, Wojciech Zaremba, Vicki Cheung, Alec Radford, and Xi~Chen.
\newblock Improved techniques for training gans.
\newblock \emph{Advances in neural information processing systems}, 29, 2016.

\bibitem[Song et~al.(2021{\natexlab{a}})Song, Meng, and Ermon]{song2020denoising}
Jiaming Song, Chenlin Meng, and Stefano Ermon.
\newblock Denoising diffusion implicit models.
\newblock \emph{arXiv preprint arXiv:2010.02502}, 2021{\natexlab{a}}.

\bibitem[Song \& Ermon(2019)Song and Ermon]{song2019generative}
Yang Song and Stefano Ermon.
\newblock Generative modeling by estimating gradients of the data distribution.
\newblock \emph{Advances in Neural Information Processing Systems}, 32, 2019.

\bibitem[Song et~al.(2021{\natexlab{b}})Song, Sohl-Dickstein, Kingma, Kumar, Ermon, and Poole]{song2020score}
Yang Song, Jascha Sohl-Dickstein, Diederik~P Kingma, Abhishek Kumar, Stefano Ermon, and Ben Poole.
\newblock Score-based generative modeling through stochastic differential equations.
\newblock \emph{arXiv preprint arXiv:2011.13456}, 2021{\natexlab{b}}.

\bibitem[Vahdat et~al.(2021)Vahdat, Kreis, and Kautz]{vahdat2021score}
Arash Vahdat, Karsten Kreis, and Jan Kautz.
\newblock Score-based generative modeling in latent space.
\newblock \emph{Advances in Neural Information Processing Systems}, 34, 2021.

\bibitem[Wang et~al.(2022)Wang, Zheng, He, Chen, and Zhou]{wang2022diffusion}
Zhendong Wang, Huangjie Zheng, Pengcheng He, Weizhu Chen, and Mingyuan Zhou.
\newblock Diffusion-gan: Training gans with diffusion.
\newblock \emph{arXiv preprint arXiv:2206.02262}, 2022.

\bibitem[Xiao et~al.(2021)Xiao, Kreis, and Vahdat]{xiao2021tackling}
Zhisheng Xiao, Karsten Kreis, and Arash Vahdat.
\newblock Tackling the generative learning trilemma with denoising diffusion gans.
\newblock \emph{arXiv preprint arXiv:2112.07804}, 2021.

\bibitem[Zhao et~al.(2020)Zhao, Liu, Lin, Zhu, and Han]{zhao2020differentiable}
Shengyu Zhao, Zhijian Liu, Ji~Lin, Jun-Yan Zhu, and Song Han.
\newblock Differentiable augmentation for data-efficient gan training.
\newblock \emph{Advances in Neural Information Processing Systems}, 33:\penalty0 7559--7570, 2020.

\end{thebibliography}
